# Tiny language models

**Ronit D. Gross[a,1], Yarden Tzach[a,1], Tal Halevi[a], Ella Koresh[a] and Ido Kanter[a,b,*]**

[a]Department of Physics, Bar-Ilan University, Ramat-Gan, 52900, Israel.
[b] Gonda Interdisciplinary Brain Research Center, Bar-Ilan University, Ramat-Gan, 52900, Israel.
*Corresponding author at: Department of Physics, Bar-Ilan University, Ramat-Gan, 52900, Israel. E-mail address: ido.kanter@biu.ac.il (I. Kanter).
[1]These authors equally contributed to this work

**Abstract**

A prominent achievement of natural language processing (NLP) is its ability to understand and generate meaningful human language. This capability relies on complex feedforward transformer block architectures pre-trained on large language models (LLMs). However, LLM pre-training is currently feasible only for a few dominant companies due to the immense computational resources required, limiting broader research participation. This creates a critical need for more accessible alternatives. In this study, we explore whether tiny language models (TLMs) exhibit the same key qualitative features of LLMs. We demonstrate that TLMs exhibit a clear performance gap between pre-trained and non-pre-trained models across classification tasks, indicating the effectiveness of pre-training, even at a tiny scale. The performance gap increases with the size of the pre-training dataset and with greater overlap between tokens in the pre-training and classification datasets. Furthermore, the classification accuracy achieved by a pre-trained deep TLM architecture can be replicated through a soft committee of multiple, independently pre-trained shallow architectures, enabling low-latency TLMs without affecting classification accuracy. Our results are based on pre-training BERT-6 and variants of BERT-1 on subsets of the Wikipedia dataset and evaluating their performance on FewRel, AGNews, and DBPedia classification tasks. Future research on TLM is expected to further illuminate the mechanisms underlying NLP, especially given that its biologically inspired models suggest that TLMs may be sufficient for children or adolescents to develop language. The data and code that support the findings of this study are openly available on https://github.com/Rg32601/Tiny-Language-Models .

1. Introduction

Two and a half centuries after the industrial revolution, which implemented routine human functionalities by machines, the artificial intelligence (AI) revolution emerged, in which machine learning can replicate many higher human functionalities[1, 2]. Beyond its practical applications, AI raises several philosophical questions, including fundamental distinctions between humans and machines.

One of the most prominent achievements of AI is its ability to understand natural language and interact efficiently with humans—a field known as natural language processing (NLP)[3, 4]. To accomplish this, NLP applies complex feedforward transformer block architectures consisting of many adaptive weights, pre-trained on massive language datasets often exceeding a terabyte, known as large language models (LLMs). This study has two primary goals.

First, it aims to determine whether pre-trained tiny language models (TLMs) exhibit features qualitatively similar to those of LLMs. TLMs are pre-trained on datasets that are $10^{-3} - 10^{-4}$ times smaller than those used for LLMs[5]; however, they cover a substantial fraction of the total token space[6, 7], which reflects their linguistic richness[8]. This question holds particular importance, as biologically inspired NLP models suggest that TLM may be sufficient for a child or adolescent to develop language.

Current LLM research requires immense computational resources accessible only to a few leading companies, far beyond the reach of most individual researchers. This study offers a positive answer to the first goal; thus, it opens new avenues for researchers to deepen their understanding of the mechanism underlying NLP and develop more efficient learning strategies with reduced computational complexity and latency.

The presented results are derived from experiments using BERT-6[9] and variants of BERT-1, optionally enhanced with additional convolutional layers (CLs) preceding the first transformer block. These models were pre-trained on a tiny subset of the Wikipedia dataset[10] comprising approximately six million paragraphs, and were subsequently fine-tuned for FewRel[11], AGNews[12] and DBPedia[13, 14] classification tasks[15].

The main results demonstrate that pre-trained TLMs can be as qualitatively efficient as LLMs, with a noticeable accuracy gap within fine-tuned classification tasks between pre-trained and non-pre-trained models. This gap increases with larger pre-training dataset sizes or greater overlap between tokens in the pre-training and classification datasets. Furthermore, the classification accuracy achieved by a deep, pre-trained TLM can be replicated by a soft committee of several shallow architectures independently pre-trained on the same TLM. This enables low-latency TLM implementations without affecting classification accuracy.

## 2. FewRel classification using pre-trained TLM on BERT-6

The FewRel dataset contains 64 output labels, each with 630 training and 70 test instances. Two strategies for pre-training BERT-6 on a subset of the Wikipedia dataset ($\sim 6 \cdot 10^6$ paragraphs) are presented.

### 2.1. *Custom-made pre-training dataset*

In the first approach, a small subset of FewRel output labels—e.g., 10 out of 64—is selected in advance. Next, a list of tokens, $T_{FR}$, comprising all tokens from their train and test instances (700 per label), is extracted. A custom subset of the Wikipedia dataset is created using only paragraphs composed solely of $T_{FR}$ tokens. This process typically reduces $W_S$ to $\sim 20{,}000 - 40{,}000$ paragraphs and limits the token set $T_W$ to $\sim 13{,}000$ compared to $30{,}522$ tokens in the full Wikipedia dataset (Table 1). A small number of missing tokens, $T_M$, may be present in $W_S$, since these $T_M$ tokens appear in paragraphs with tokens out of $T_{FR}$ in the entire Wikipedia dataset. This pre-training custom made dataset offers a straightforward way to reduce the embedding dimension, lowering memory and computational requirements.

BERT-6 is then pre-trained on the custom-made $W_S$ and fine-tuned (via transfer learning[16, 17]) on the FewRel training dataset ($10 \times 630$), with performance evaluated on the corresponding test dataset ($10 \times 70$). Variability in $T_W, W_S,$ and final accuracy is dependant on the selected 10 output labels among the 64 (Table 1). Additional fluctuations arise from variations in the accuracy per label after fine-tuning BERT-6, which is pre-trained on the entire Wikipedia dataset (Fig. 1)[9]. Nevertheless, the results reveal a consistent accuracy gap of ~0.06 between FewRel test performance with and without pre-training BERT-6 on $W_S$ (Table 1). This gap is smaller than the ~0.11 gap observed when pre-training BERT-6 on the full Wikipedia dataset; however, it still demonstrates the efficiency of the TLM-based pre-training approach.

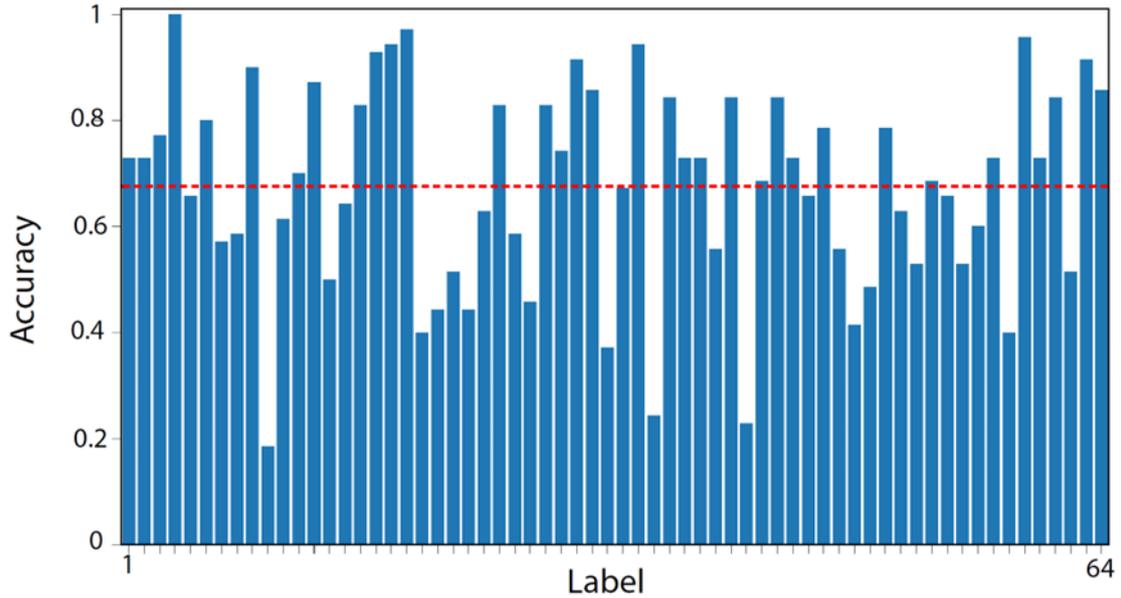

**Fig. 1.** Accuracy per label. Pre-trained BERT-6 on the full Wikipedia dataset, followed by fine-tuning on the complete FewRel training dataset ($64 \times 630$). Accuracy is evaluated using the test dataset (70 instances per label), revealing high variance around the average accuracy of $0.676$ (indicated by the dashed red horizontal line).

| FewRel with 10 output labels | | | | |
|---|---|---|---|---|
| $Ws$ | $T_W$ | $T_M$ | $Acc.$ | $Gap$ |
| 37,936 | 12,829 | 1,053 | 0.864 | 0.066 |
| 20,589 | 13,959 | 6,492 | 0.795 | 0.069 |
| 46,898 | 13,342 | 1,084 | 0.871 | 0.058 |

**Table 1.** Accuracy results for BERT-6 pre-trained on the custom-made $W_S$ dataset and fine-tuned on 10 selected FewRel output labels ($10 \times 630$ training,

$10 \times 70$ testing). Reported metrics include the total number of tokens in the FewRel training and test datasets, $T_{FR}$, the number of missing tokens $T_M$ among $T_{FR}$ in $W_S$, final test accuracy, $Acc.$, and the observed accuracy gap in the $Acc.$ with and without pre-training, $Gap$. The selected label sets for each rows are [5, 10, 11, 16, 17, 25, 27, 33, 36, 39], [8, 13, 18, 23, 26, 30, 42, 49, 50, 56], and [0, 2, 5, 10, 11, 16, 17, 36, 45, 63], respectively.

The improved achieved accuracy, as indicated by the gap, is not limited to a small subset of labels (Table 1); similar improvements were observed across all 64 FewRel output labels. This was achieved by constructing a compact pre-training dataset $W_S$, using a limited number of training and test examples per label, thereby constraining the overall token set (Table 2). For example, selecting only 20 train/test instances per each of the 64 labels resulted in a smaller token set $T_{FR}\sim 11{,}000$, while the size of $W_S$ increased to ~60,000 and the observed accuracy gap $Gap$ increased to ~0.15, significantly higher than that reported in Table 1.

| | | FewRel with 64 output labels | | | | |
|---|---|---|---|---|---|---|
| $Train_{size}$ | $Test_{size}$ | $Ws$ | $T_{FR}$ | $T_M$ | Acc. | Gap |
| 20 | 20 | 60,655 | 11,135 | 764 | 0.456 | 0.149 |
| 30 | 20 | 105,101 | 12,306 | 562 | 0.476 | 0.123 |

**Table 2.** Accuracy results for BERT-6 pre-trained on the $W_S$ custom-made dataset using all 64 FewRel output labels (similar to Table 1 with 10 labels). $Train_{size}/Test_{size}$ denotes train/test dataset sizes, respectively. Other notations are as defined in Table 1.

### 2.2. Randomly selected pre-training dataset

In the custom-made TLM scenario, $W_S$ was constructed using $T_{FR}$. Here, similar results were obtained by pre-training BERT-6 on randomly selected small subsets of $W_S$, which cover nearly the entire token set of Wikipedia ($T_w = 30{,}522$), followed by fine-tuning on the FewRel training dataset. Results indicate that the accuracy gap decreases as $W_S$ decreases, from 0.07 for $W_S = 90,000$ to ~0.01 for $W_S = 2,000$ (Table 3). The reduction in $T_W$ from $W_S = 90,000$ to 2,000 is modest; however, the frequency of appearance of each token drops significantly, contributing to the observed decrease in both accuracy and the gap. Remarkably, even $W_S = 90,000$, which is only about 1.5% of the entire Wikipedia dataset, achieves a gap equivalent to ~75% of the maximum observed gap (Table 3). Additionally, for $W_S \in [20,000, 40,000]$, the observed gap $[0.043, 0.05]$ (Table 3a) is slightly smaller than the 0.06 which reported for similar $W_S$ sizes in Table 1. This difference may stem from pre-training on smaller tokens $T_{FR}$ set (Table 1), compared to the full Wikipedia token set (30,552) (Table 3).

The decrease in accuracy and the corresponding gap is associated with two simultaneous trends (Table 3): a decrease in $W_S$ size and an increase in the number of missing tokens $T_M$, which measures the dissimilarity between the tokens in $W_S$ and those in the FewRel dataset. These factors must be decoupled to determine whether each of them independently contributes to a decrease in accuracy. $T_M$ remains nearly constant across the first two rows of Table 3; therefore, the observed decrease in accuracy is attributed to the substantial reduction in $W_S$ size. Accuracy must be compared under the same $W_S$ and $T_{FR}$, but with different values of $T_M$ to verify whether $T_M$ solely affects accuracy. This is achieved by using a fixed $W_S$ from Tables 1 or 3 but artificially increasing $T_M$ by excluding a substantial portion of $T_{FR}$ (Table 4 and Appendix). This deliberate mismatch between $T_W$ and $T_{FR}$ sets resulted in a considerable decrease in accuracy—comparable to the accuracy achieved at $W_S = 5,000$ (Table 4)—and produced a similarly small gap (Table 3). The results underscore the importance of token overlap and alignment between the pre-training and classification datasets. Increasing the size of the randomly selected $W_S$ improves this similarity and raises the token frequency of each token, further enhancing classification accuracy.

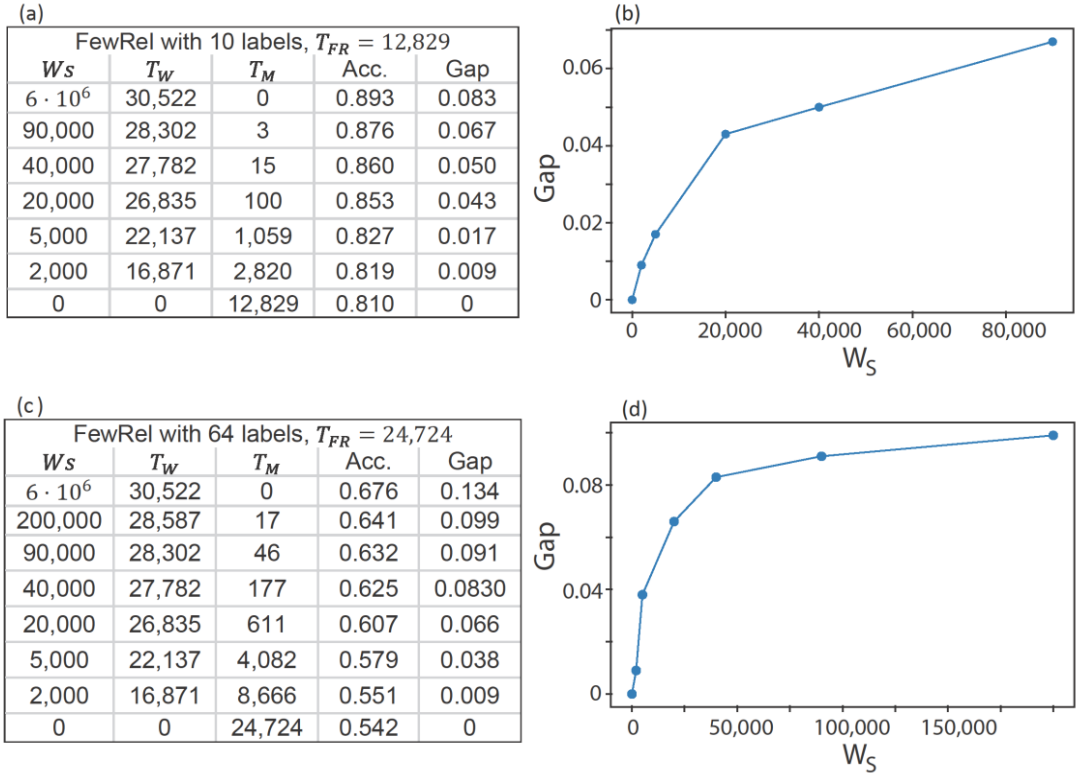

**Table 3.** (a) Pre-training BERT-6 on randomly selected Wikipedia subsets of varying size $W_s$, each consisting of $T_W$ tokens. Fine-tuning is performed on FewRel with 10 output labels using 630/70 train/test instances and 12,829 unique tokens. $T_M$ denotes the number of tokens in the FewRel datasets not present in $W_s$. Accuracy, $Acc.$, and the accuracy $Gap$ with and without pre-training are reported. (b) $Gap$ values from panel (a) as a function of $W_S$. (c-d) Similar to panels (a) and (b), but for 64 output labels, each with 630/70 train/test FewRel instances.

(a)

| FewRel with 10 output labels, $T_{FR} = 12,829$ | | | |
|---|---|---|---|
| $W_S$ | $T_W$ | $T_M$ | Acc. |
| 40,000 | 27,782 | 15 | 0.860 |
| 40,000 | 14,799 | 8,661 | 0.823 |
| 5,000 | 22,137 | 11,059 | 0.827 |

(b)

| | FewRel with 64 output labels, $T_{FR} = 24{,}724$ | | |
|---|---|---|---|
| $W_S$ | $T_W$ | $T_M$ | Acc. |
| 40,000 | 27,782 | 177 | 0.625 |
| 40,000 | 9,658 | 16,301 | 0.578 |
| 5,000 | 22,137 | 4,082 | 0.579 |

**Table 4.** (a) FewRel with 10 labels and $T_{FR} = 12{,}829$, pre-trained on $W_S = 40{,}000$, where $T_M$ is artificially increased (middle row, highlighted in light blue). The first and third rows are included from Table 3(a) for comparison. Accuracy in the high $T_M$ case (second row) is similar to that for $W_S = 5{,}000$ (third row). (b) Similar to panel (a), but for FewRel with 64 labels. The first and third rows are taken from Table 3(b).

### 3. AGNews classification using pre-trained TLM on BERT-6

Similar results to those obtained for FewRel fine-tuning (Section 2.2) were obtained for the AGNews dataset, which consists of four output labels (Table 5). The full AGNews train and test datasets yield near-perfect accuracy (~0.99), with an observed gap below 0.03 even for small $W_S$; therefore, results are reported for a reduced dataset of 1,000 train and 1,000 test instances per label, comprising $T_{AGN} = 17{,}072$ tokens (Table 5). The results reveal that the accuracy gap decreases with smaller $W_S$ whereas $T_M$ increases, similar to the trends observed in FewRel (Table 3). Notably, even at $W_S = 2{,}000$, the gap remained evident at 0.014, demonstrating the effectiveness of TLM pre-training for AGNews (Table 5).

| AGNews |||||
| $T_{AGN} = 17{,}072$, 1,000 train/test instances per label |||||
| $W_s$ | $T_W$ | $T_M$ | Acc. | Gap |
|---|---|---|---|---|
| $6 \cdot 10^6$ | 30,522 | 0 | 0.935 | 0.095 |
| 90,000 | 28,302 | 90 | 0.891 | 0.051 |
| 40,000 | 27,798 | 249 | 0.878 | 0.038 |
| 5,000 | 22,137 | 2,736 | 0.861 | 0.021 |
| 2,000 | 16,871 | 5,473 | 0.854 | 0.014 |
| 0 | 0 | 17,072 | 0.840 | 0 |

**Table 5.** BERT-6 pre-trained on a randomly selected Wikipedia subset of size $W_s$, consisting of $T_W$ tokens, followed by fine-tuning on a limited AGNews dataset (1,000 train/test instances per label). Notations follow those in Table 3.

## 4. DBPedia classification using pre-trained TLM on BERT-6

The DBPedia dataset consists of 14 output labels, with 40,000/5,000 train/test instances per label, comprising $T_{DBP} = 28{,}621$ unique tokens. When using the full dataset, the accuracy with and without pre-training is ~0.99, with a minimal gap of <0.01 [18]. To observe a more meaningful gap, the dataset size was significantly reduced to 100 train/test instances per label, yielding $T_{DBP} = 15{,}084$ (Table 6). The accuracy remained high; however, the gap between pre-trained and non-pre-trained models increased with $W_S$, reaching up to 0.053, while $T_M$ decreased (Table 6), highlighting the effectiveness of TLM pre-training for DBPedia classification.

Increasing $T_M$ while keeping $W_S$ and $T_{DBP}$ constant led to decreased accuracy (Table 7), consistent with the pattern observed in FewRel (Table 4). Specifically, for $W_S = 20{,}000$, artificially increasing $T_M$ reduced accuracy to below that of the randomly selected $W_S = 5{,}000$, and close to the accuracy achieved without pre-training (Table 7).

| DBPedia | | | | |
|---|---|---|---|---|
| $T_{DBP} = 15{,}084$, 100 train/test instances per label | | | | |
| $W_S$ | $T_W$ | $T_M$ | Acc. | Gap |
| $6 \cdot 10^6$ | 30,522 | 0 | 0.991 | 0.053 |
| 90,000 | 28,302 | 37 | 0.984 | 0.047 |
| 20,000 | 26,835 | 249 | 0.979 | 0.041 |
| 10,000 | 25,089 | 620 | 0.972 | 0.034 |
| 5,000 | 22,137 | 1,470 | 0.957 | 0.019 |
| 0 | 0 | 15,084 | 0.938 | 0 |

**Table 6.** BERT-6 pre-trained on randomly selected Wikipedia subsets $W_S$ with corresponding $T_W$ tokens, followed by fine-tuning on a limited DBPedia dataset (100 train/test instances per label). Notations are consistent with those in Table 3.

| DBPedia, $T_{DBP} = 15{,}084$ | | | |
|---|---|---|---|
| $W_S$ | $T_W$ | $T_M$ | Acc. |
| 20,000 | 26,835 | 249 | 0.979 |
| 20,000 | 12,006 | 10,250 | 0.948 |
| 5,000 | 22,137 | 1,470 | 0.957 |
| 0 | 0 | 15,084 | 0.938 |

**Table 7.** DBPedia classification with a reduced train/test dataset ($T_{DBP} = 15{,}084$, as in Table 6). Pre-training was performed on $W_S = 20{,}000$. The second row (highlighted in light blue) reflects an enhanced $T_M = 10{,}250$, compared to $T_M = 249$ in the first row for a randomly selected $W_S$. For comparison, the first, third, and fourth rows are replicated from Table 6 for randomly selected $W_S$. The achieved accuracy for the high $T_M$ case (second row) falls between the results for $W_S = 5{,}000$ and $W_S = 0$ (no pre-training).

## 5. TLMs with low latency

Pre-training BERT-6 on FewRel (10 output labels) with $W_S \sim 40,000$ and $T_{FR} \sim 13,000$ tokens yields an accuracy of $\sim 0.865$ and a gap of $\sim 0.06$ (Tables 1 and 8). In comparison, BERT-1—containing only one transformer block—achieves $0.841$, $0.024$ below BERT-6.

Previous results on the compact convolutional transformer (CCT-7) indicated that different vision transformer (ViT) architectures[19] with comparable accuracies differ in the properties of their single-head performance[20, 21]. Thus, combining several shallow variants with a soft-committee vote improved accuracy[22] to equal that of a much deeper CCT-7 model[23]. This gain in accuracy (Appendix) surpassed the improvement obtained by ensembling identical architectures trained from different random initial conditions. These shallow ViT architectures differ in the number of multi-head attention (MHA) heads[4, 21], MHA dimensions, and the inclusion of a CL before the first transformer block.

Similar effects were obtained for NLP, using five pre-trained BERT-1 variants that varied in number of heads, MHA dimension, the number of preceding CLs and their filters, then independently fine-tuned on FewRel [24, 25] (Table 8). Their soft-committee output reached $0.866$ accuracy—identical to that achieved by a single BERT-6 model (Table 8)[9]. The total number of layers of an architecture correlates directly with network latency; therefore, the latency of BERT-6 is $25$ $(6 \times 4 + 1)$, whereas that of BERT-1 with two CLs (Table 6) is only $7$ $(4 + 3)$. The pattern holds as well for the full 64‑label FewRel task: the ensemble of five BERT‑1 variants attains $0.634$ accuracy, exceeding the $0.625$ accuracy of a single BERT-6 (Table 8). These findings demonstrate that a soft committee of several shallow, low-latency architectures can match or surpass the accuracy of a single, substantially deeper model, while delivering significantly lower inference latency.

The use of 2-dimensional kernels before the BERT-1 transformer block (Table 8) further enhanced accuracy compared to 1-dimensional kernels applied only along the $768$ embedding dimension of each token[3, 26, 27]. These types of CLs function like sliding windows along the embedding dimension and the token sequence, but are not designed to cover the entire receptive field ($128 \times 768$) as in a standard Convolutional Neural Networks (CNNs) [28-31].

For the AGNews dataset, the effect of a soft committee is further enhanced. A soft committee composed of only three independently trained BERT-1 models—each initialized with random weights and without any pre-training—achieved an accuracy of $0.868$(Table 9), surpassing the $0.861$ accuracy of BERT-6 pre-trained on $W_S = 5,000$ (Table 4). This indicates that pre-training may not be necessary for achieving high performance on simpler tasks, and that ensembling a shallow, non-pre-trained BERT-1 architecture is sufficient to reach or exceed the performance accuracy of deeper pre-trained models. This result may stem from the relative simplicity of the AGNews classification task, which achieves very high accuracy on the full training and test sets and maintains relatively high accuracy even on substantially reduced datasets (Table 5).

(a)

| FewRel with 10 labels $W_S = 37,936$, $T_{FR} = 12,829$, Acc. = 0.866 | | | | | |
|---|---|---|---|---|---|
| Bert blocks | Convolution layers | $d$ | kernel | No. heads | Acc. |
| 1 | 2 | 64 | $3 \times 3$ | 8 | 0.846 |
| 1 | 2 | 64 | $3 \times 3$ | 12 | 0.841 |
| 1 | 0 | - | - | 24 | 0.841 |
| 1 | 2 | 64 | $16 \times 3$ | 12 | 0.836 |
| 1 | 3 | 32 | $3 \times 3$ | 12 | 0.836 |

(b)

| FewRel with 64 labels. $W_S = 40,000$, $T_{FR} = 24,724$, Acc. = 0.634 | | | | | |
|---|---|---|---|---|---|
| Bert blocks | Convolution layers | $d$ | kernel | No. heads | Acc. |
| 1 | 2 | 64 | $3 \times 3$ | 8 | 0.592 |
| 1 | 2 | 64 | $3 \times 3$ | 12 | 0.607 |
| 1 | 0 | - | - | 24 | 0.592 |
| 1 | 2 | 64 | $16 \times 3$ | 12 | 0.614 |
| 1 | 3 | 32 | $3 \times 3$ | 12 | 0.610 |

**Table 8.** (a) Five BERT-1 models pre-trained on $W_S$ (with token set $T_{FR}$ described in section 2.1), incorporating optional CLs before the single transformer block. The transformer dimension is defined as $64 \times No.\,heads$. Upper table: Fine-tuning on FewRel with 10 output labels (as in Table 1) yields an average accuracy of 0.84 across the five models. Their soft committee achieves 0.866 accuracy, matching that of BERT-6 pre-trained on the same dataset (Table 1). Lower table: For FewRel with 64 output labels, the five BERT-1 models achieve 0.603 average accuracy for the five architectures, and their soft committee reaches 0.634, exceeding the 0.625 accuracy of BERT-6 (Table 3).

| AGNews | | | | | |
|---|---|---|---|---|---|
| $T_{AGN} = 17{,}072,\ Committee\ Acc. = 0.868$ | | | | | |
| BERT blocks | CLs | $d$ | kernel | No. heads | Acc. |
| 1 | 0 | - | - | 12 | 0.839 |
| 1 | 0 | - | - | 12 | 0.834 |
| 1 | 1 | 64 | $3 \times 3$ | 12 | 0.820 |

**Table 9.** Two variants of BERT-1, each initialized with random weights and trained without pre-training, were fine-tuned on the AGNews dataset using 1,000 train and test instances per label (as in Table 5). The soft committee of the three trained architectures achieved an accuracy of 0.868, exceeding the 0.861 accuracy of BERT-6 pre-trained on $W_S = 5{,}000$ (Table 5).

## 6. Discussion

Pre-training TLMs[32, 33] demonstrates qualitative efficiency comparable to that of LLMs, as evidenced by the fine-tuning results across several classification tasks. These tasks consistently reveal a gap in accuracy between models with and without pre-training. This gap increases rapidly with the initial growth in pre-training dataset sizes, and at $W_S = 90{,}000$—which represents only 1.5% of the full Wikipedia dataset—models achieve 50% to 90% of the maximal possible gap (Tables 1-7). This suggests that the near maximal gap is achieved for much smaller datasets, offering a substantial reduction in $time \times space$ complexity of the pre-training process without compromising classification accuracy. A comparison across different classification tasks reveals considerable variation in the dataset size $W_S$ required to achieve a given fraction of the maximum gap—an observation that opens avenues for further investigation.

A key factor influencing TLM performance is the overlap between tokens in the pre-training and classification datasets. As this overlap decreases, pre-training becomes less efficient, and classification accuracy decreases, as intuitively expected. Artificially increasing the number of missing tokens $T_M$ in $W_S$ relative to those required by the classification dataset reduces accuracy towards levels observed without pre-training. Given the substantial variability in token frequency within datasets like Wikipedia, it would be interesting to generalize the concept of overlap into a weighted overlap, which accounts for the token appearance frequency. For example, a small subset of high-frequency tokens (e.g., 1000) appears frequently, while most tokens occur infrequently; similar trends emerge in the classification dataset. However, the impact of this frequency-weighted overlap on pre-training efficiency remains unclear.

The results highlight the feasibility of achieving low-latency TLMs through soft committee ensembles of shallow architectures, achieving comparable accuracy to much deeper models[34]. This low latency is an important feature for real-life implementations. Notably, the space complexity of five BERT-1 shallow variants is similar to—or not significantly greater than—that of a single BERT-6 model (Table 7) [35]. Nevertheless, the results are based on a limited number of examples, and further research is needed to determine the maximum achievable accuracy with larger ensembles and to identify optimal shallow architecture configurations.

The presented results on TLMs open an opportunity to examine the interplay among four fundamental quantities: the pre-training dataset size, $W_S$, composed of $T_W$ tokens, and the classification dataset size, $W_C$, composed of

$T_C$ tokens. A large $W_S$ increases $T_W$, both of which increase the pre-training $time \times space$ complexity. Moreover, a significant mismatch between $T_W$ and $T_c$ may introduce noise into the classification task due to token irrelevance. The findings highlight the importance of minimizing $T_M$, the number of missing tokens in $T_C$. However, the effect of the reverse case (tokens in $W_S$ absent from $T_C$) has not been fully explored. High frequency tokens in $T_M$ may contribute disproportionately to the pre-training and reduce its efficiency. Hence, merely enlarging the dataset space ($W_S, T_W, W_C, T_C$) does not necessarily improve accuracy. Instead, the correlation between pre-training and classification datasets plays a crucial role. A rational conclusion is that using smaller, task-specific pre-training datasets can improve performance and reduce computational cost. However, this approach requires prior knowledge of the domain or query space.

Life experiences suggest that TLM-level exposure may be sufficient for a child or adolescent to develop a language, and in many circumstances, for adults to quickly become familiar with a new language or a research area. The presented NLP architectures, along with their pre-training and fine-tuning processes, differ significantly from the biological realities of brain activity[36]; however, the successful implementation of NLP is a shared goal across both disciplines. Some of the learning principles of effective TLMs will likely be relevant for understanding the underlying mechanisms of brain-inspired NLP.

TLMs also offer individual research groups the opportunity to explore and deepen their understanding of NLP implementation, which is currently not feasible for LLMs due to their massive computational requirements. The direct applicability of the presented results to LLMs remains uncertain; however, they provide a valuable foundation for further investigation and extension by organizations with the necessary resources. The mechanism underlying successful deep learning, such as matrix-based performance metrics for CNN filters[37, 38] and single-head performance [23] metrics for ViT architectures[19-21], could be generalized to TLMs. A better understanding of the pre-training process, the role of embedding dimension, and the specific functions of individual heads within the MHA and across transformer blocks may inform more efficient pruning strategies and the development of optimized learning architectures.

The presented perspective on NLP, particularly the quantitative analysis of TLMs, already raises several open questions and promising research directions that are expected to drive further advancement in the field.

**Acknowledgements**

The work is supported by the Israel Science Foundation [grant number 346/22]

**Appendix**

*1. Dataset and preprocessing*

The datasets used in this study are DBpedia[13], AGNews[12], FewRel[11] and Wikipedia[10]. Each dataset was tokenized using the BERT tokenizer from the HuggingFace Transformers library[8], specifically the bert-base-uncased variant[9], which converts raw text into $30,522$ token IDs. Tokenization was performed using the following configuration: truncation to a maximum length of 128 tokens and padding to the same length.

*2. Optimization*

The CrossEntropyLoss[39] function was selected for the classification task and minimized using the stochastic gradient descent algorithm[40, 41] and the AdamW optimizer[42] was used. The maximal accuracy was determined by searching through the hyper-parameters (see below). The L2 regularization method[43] was applied.

To pre-train our models, we employed a Masked Language Modeling (MLM) objective similar to that used in the original BERT architecture[9]. Pre-training was performed on a filtered Wikipedia-derived dataset. The MLM procedure followed the standard masking strategy[8]: 15% of tokens were selected for masking, of which 80% were replaced with [MASK], 10% with random tokens, and 10% remained unchanged. Special tokens (e.g., [PAD], [CLS], [SEP]) were excluded from masking.

3. *Hyper-parameters*

The hyper-parameters $\eta$ (learning rate) and $\alpha$ (L2 regularization) were optimized for offline learning, using a mini-batch size of 32 inputs. The learning-rate decay schedule was also optimized. For all the simulations, a linear scheduler was applied using the HuggingFaceutility[8], without warm-up steps and the schedular was updated only for the first 50 epochs. This schedule gradually decays the learning rate from its initial value to zero in a linear fashion throughout training, which helps stabilize convergence[44]. The pre-training models in all the simulations were trained for 50 epochs with $\eta = 5.5e - 5, \alpha = 1e - 2$ and a linear scheduler was applied. The fine-tuning process was trained for at least 50 epochs.

Statistics for each data point were computed based on at least five trials, with fluctuations being around 1%.

For Fig.1, we utilized the pre-trained DistilBERT model using the HuggingFace Transformers library[45, 46] and fine-tuned it for the 64 FewRel label classification task using $\eta = 1e - 4, \alpha = 1e - 2$.

For Table 1, we pre-trained a BERT-6 model on a custom Wikipedia-based dataset, consisting of all paragraphs that contained only tokens appearing in the 10-label subset of the FewRel dataset. Following pre-training, the model was fine-tuned specifically to classify those 10 output labels with $\eta = 8e - 5, \alpha = 1e - 2$. Without pre-training, the model was randomly initialized and directly fine-tuned to classify the 10 target labels using $\eta = 8e - 5, \alpha = 1e - 2$.

For the first row in Table 2, we pre-trained a BERT-6 model on a custom Wikipedia-based dataset, all paragraphs composed solely of tokens found in selected examples from the FewRel dataset. Following pre-training, the model was fine-tuned specifically to classify the FewRel 64 output labels with $\eta = 5e - 5, \alpha = 1.1e - 2$. The model without pre-training was randomly initialized and directly fine-tuned on the same classification task using identical hyperparameters. In the second row of Table 2, the procedure remained the same, but with the following hyperparameters: the pre-trained model was fine-tuned with $\eta = 5e-5, \alpha = 1.1e-2$ while the non-pre-trained model was fine-tuned with $\eta = 1e-5, \alpha = 1.1e-2$.

For Table 3, for $W_s = 6 \cdot 10^6$ we utilized the pre-trained DistilBERT model and fine-tuned it for the 10 FewRel label classification task. For other values of $W_s$ we pre-trained a BERT6 model from scratch on a Wikipedia-derived dataset composed of $W_s$ randomly selected paragraphs, using a fixed random seed of 42 to ensure reproducibility. The hyper parameters used for panels (A) and (B) are presented in Table 10:

| $W_s$ | $\eta$ | $\alpha$ |
|---|---|---|
| $6 \cdot 10^6$ | 5e-5 | 1.2e-2 |
| 90,000 | 5e-5 | 1e-3 |
| 40,000 | 5e-5 | 2e-2 |
| 20,000 | 7e-5 | 1.5e-2 |
| 5,000 | 5e-5 | 1.5e-2 |
| 2,000 | 5e-5 | 5e-3 |
| 0 | 5e-5 | 1.2e-2 |

**Table 10.** Hyper-parameters for Table 3(a) and 3(b).

The hyper parameters used for panels (c) and (d) in Table 3 are presented in Table 11:

| $W_s$ | $\eta$ | $\alpha$ |
|---|---|---|
| $6 \cdot 10^6$ | 5e-5 | 1.2e-2 |
| 200,000 | 5e-5 | 1.2e-2 |
| 90,000 | 5e-5 | 1.2e-2 |
| 40,000 | 5e-5 | 1.5e-2 |
| 20,000 | 5e-5 | 1e-3 |
| 5,000 | 5e-5 | 1e-2 |
| 2,000 | 5e-5 | 1e-2 |
| 0 | 5e-5 | 5e-3 |

**Table 11.** Hyper-parameters for Table 3(c) and (d).

For Table 4, the pre-training Wikipedia dataset was constructed by first identifying tokens not present in the FewRel dataset and then extending this list with additional likely tokens from the Wikipedia corpus. The Wikipedia dataset was subsequently filtered to include only paragraphs that contained only these selected tokens, resulting in a pre-training corpus tailored to emphasize complementary information. The hyper-parameters used are $\eta = 5.5e - 5, \alpha = 2e - 2$.

For Table 5, similar to Table 3 for $W_s = 6 \cdot 10^6$ we utilized the pre-trained DistilBERT model and fine-tuned it for the 4 AGNews label classification task. For other values of $W_s$ we pre-trained we pre-trained a BERT6 model from scratch on a Wikipedia-derived dataset composed of $W_s$ randomly selected paragraphs, using a fixed random seed of 42 to ensure reproducibility. The hyper parameters are presented in Table 12:

| $W_s$ | $\eta$ | $\alpha$ |
|---|---|---|
| $6 \cdot 10^6$ | 5e-5 | 1.2e-2 |
| 90,000 | 5e-5 | 1.2e-2 |
| 40,000 | 5e-5 | 1.5e-2 |
| 20,000 | 5e-5 | 1e-3 |
| 5,000 | 5e-5 | 1e-2 |
| 2,000 | 2e-4 | 8e-2 |
| 0 | 5e-5 | 5e-3 |

**Table 12.** Hyper-parameters for Table 5.

For Table 6, similar to Table 3 for $W_s = 6 \cdot 10^6$ we utilized the pre-trained DistilBERT model and fine-tuned it for the 14 DBPedia label classification task. For other values of $W_s$ we pre-trained we pre-trained a BERT6 model from scratch on a Wikipedia-derived dataset composed of $W_s$ randomly selected paragraphs, using a fixed random seed of 42 to ensure reproducibility. The hyper parameters are presented in Table 12:

For Table 7, Similar to Table 4, the pre-training Wikipedia dataset was constructed by first identifying tokens not present in the DBPedia dataset and then extending this list with additional most common likely tokens from the Wikipedia corpus. The Wikipedia dataset was subsequently filtered to include only paragraphs that contained only these selected tokens, resulting in a pre-training corpus tailored to emphasize complementary information. The hyper-parameters used are $\eta = 5e-5, \alpha = 1.2e-2$.

For Table 8, the hyperparameters used for the fine-tuning in both panels (a) and (b) are $\eta = 1e-4, \alpha = 1e-2$. The pre-training datasets used in Tables 8 panels (a) and (b) correspond to those used in the colored rows of Table 3 panels (a) and (b), respectively.

For Table 9, The hyper-parameters are $\eta = 1e-4, \alpha = 1.1e-2$.

*2.4. Statistics*

Statistics for all results were obtained using at least five samples and the standard division was around $1\%$ for all the results.

*2.5. Soft Committee*

The soft committee decision was performed by the summation of all output fields without any alterations such as Softmax, activation or normalization. The decision was then made on the summed field.

*2.6. Hardware and software*

We used Google Colab Pro and its available GPUs. We used Pytorch for all the programming processes.